\documentclass{article}

\usepackage{PRIMEarxiv}
\usepackage[utf8]{inputenc}
\usepackage{amsmath}
\usepackage{amsfonts}
\usepackage{amssymb}
\usepackage{mathtools}
\usepackage{booktabs}
\usepackage{graphicx}
\usepackage{microtype}
\usepackage{nicefrac}
\usepackage[numbers,sort&compress]{natbib}
\usepackage{url}
\usepackage{hyperref}

\graphicspath{{figures/}}
\hypersetup{
  colorlinks=true,
  citecolor=blue,
  linkcolor=blue,
  urlcolor=blue
}

\AtBeginDocument{%
  }

\begin{document}

\title{MixFormer: Linear Transformer with Mixture of Memory Experts}

\author{
  Yu Guo \\
  \normalfont School of Computer Science, Sichuan University \\
  Chengdu, China \\
  \texttt{gyguoyugy@gmail.com}
  \And
  Lei Duan\thanks{Corresponding author.} \\
  \normalfont School of Artificial Intelligence, Sichuan University \\
  Chengdu, China \\
  \texttt{leiduan@scu.edu.cn}
}

\maketitle

\begin{abstract}




State Space Models (SSMs), as a mainstream research direction of linear Transformers, aim to achieve higher efficiency than standard Transformers in long-context modeling.
However, existing SSMs suffer from limited input adaptivity and constrained memory capacity, leading to information loss when modeling ultra-long sequences.
To address these limitations, we propose MixFormer, a novel linear Transformer that integrates a Mixture-of-Memory-Experts (MoE) mechanism.
Specifically, the model maintains differentiated memory states through multiple collaborating memory experts and employs a novel Time-Aware Linear Attention (TALA) mechanism, which leverages learnable exponential decay functions and positional biases to dynamically update memory.
This design enables the model to selectively reinforce important historical information while effectively mitigating memory dilution, substantially improving long-range dependency modeling.
Experiments on long-sequence text and image generation tasks demonstrate that MixFormer not only achieves significant performance gains but also provides a more sustainable computational backbone for the next generation of web infrastructure.
\end{abstract}





\section{Introduction}





Transformer \cite{vaswani2017attention} has become the foundational architecture for large language models (LLMs) \cite{touvron2023llama,touvron2023llama2} and multimodal models \cite{dubey2024llama,deitke2024molmo,dai2024nvlm}, having been successfully applied in fields such as Natural Language Processing (NLP) and Computer Vision (CV).
The powerful representation learning capability of Transformers stems from the self-attention mechanism, which leverages the softmax function to capture full pairwise token interactions.
This results in quadratic time and space complexity ($O(L^2)$) with respect to the context length, making it unsuitable for very long-range sequence modeling.
In the World Wide Web (WWW), the Transformer architecture has deeply penetrated a wide range of core applications — from semantic understanding in search engines and personalized generation in recommendation systems to real-time interactive processing in web browsers. Its self-attention–based parallel computation offers highly efficient sequence modeling capabilities for massive web-scale data.

Recently, efficient Transformer models \cite{tay2022efficienttransformerssurvey, zhuang2023survey,papa2024survey,han2022survey} have attracted increasing research interest, with a primary focus on reducing the computational cost of the self-attention mechanism.
Among these efforts, linear Transformers have emerged as one of the mainstream solutions for building efficient Transformers.
The core module of linear Transformers lies in constructing linear attention mechanisms that approximate the behavior of the softmax function through carefully designed kernel functions, thereby reducing the computational cost from $O(N^2)$ to $O(N)$, where $N$ is the context length.

In addition, other linear Transformers further extend linear attention into a recursive formulation, allowing each token at a given time step to inherit the computation results from the previous step.
Such methods are typically categorized as State Space Models (SSMs) \cite{guefficiently, gupta2022diagonal, gu2021combining, hasaniliquid, smith2023simplified}.
Unlike the self-attention mechanism, which requires computing an explicit attention score matrix, SSMs do not maintain the correspondence between each query and historical tokens when handling variable-length contexts.
Instead, an SSM preserves a fixed-size intermediate feature map as a global historical memory state matrix, which encodes and compresses information from all previous time steps while discarding their explicit temporal indices.
Specifically, during the recursive process at each time step, the SSM's linear attention updates the memory state matrix through additive operations.
This update strategy discards the positional indexing of the original queries, making it impossible to retrieve information from specific time steps.

Existing approaches primarily enhance the memory capability of SSM to capture historical information by introducing RNN-style gating mechanisms. 
These methods employ either fixed decay coefficients (e.g., the constant parameter proposed by \cite{sun2023retentive}) or learnable weight matrices (e.g., the work of \cite{yang2024gated}), enabling the model to retain past states to some extent.
However, although such methods can marginally increase the capacity of the SSM’s memory state, they still suffer from the following inherent limitations:
\textbf{(1)}
Existing approaches typically update the memory state using fixed coefficients or a limited number of learnable parameters, resulting in insufficient input adaptivity. 
The model should dynamically decide, at each time step, which information to preserve over the long term and which to gradually decay based on its contextual importance.
\textbf{(2)}
Most current SSMs employ a single memory state matrix to store historical information. Under ultra-long context inputs, this single-state architecture inherently limits storage capacity, making it difficult to prevent early critical information from being overwritten or diluted by subsequent inputs, thereby leading to irreversible information loss in long-range dependency modeling.

To bridge these gaps, we propose a novel linear Transformer with Mixture of Memory Experts (\textbf{MixFormer}).
The proposed architecture is built upon the linear Transformer, supporting both parallel training and recurrent inference modes.
Specifically, MixFormer introduces the Mixture-of-Experts (MoE) into state memory modeling for the first time, where multiple memory experts collaboratively construct the final memory state through weighted fusion.
Within each expert, we design a novel Time-Aware Linear Attention (TALA) to replace the conventional multi-head attention layer.
By integrating exponential decay with learnable parameters, TALA dynamically adjusts the memory update strategy according to the current input, enabling the model to reinforce important historical information while progressively attenuating less relevant content.
Furthermore, the multi-expert parallel architecture breaks the capacity limitation of a single memory state by forming a distributed memory storage.
Each expert employs a differentiated memory strategy via the TALA mechanism, focusing on distinct temporal patterns such as long-range dependencies or short-term dynamics.
Through a content-adaptive expert activation mechanism, MixFormer selectively strengthens the preservation of salient information and suppresses redundant signals.      
After multi-layer stacking, the model achieves a robust capability for ultra-long context modeling.

We conduct extensive experiments on both image generation (including conditional and unconditional generation) and long-sequence text modeling tasks. 
The results demonstrate that the proposed MixFormer consistently achieves significant performance gains across all tasks.
Furthermore, the effectiveness of MixFormer is validated through an in-depth analysis of its causal masking mechanism and the collaborative division of multiple memory experts.
Finally, with the growing demand for web applications that support long-text understanding and multimodal content processing, MixFormer is expected to provide a more sustainable computational backbone for the next generation of web infrastructure.
The main contributions are as follows:
\begin{itemize}
    \item 
    We propose a novel linear Transformer with mixture-of-memory-experts (MixFormer). 
    To the best of our knowledge, this is the first work to incorporate multi-memory-experts into State Space Models (SSMs). 
    Maintaining multiple differentiated memory states in parallel enhances global perception over ultra-long sequences.
    \item 
    We propose a Time-Aware Linear Attention (TALA) mechanism based on linear attention, which integrates an exponential decay function with learnable pair-wise position biases to significantly enhance the model’s capability for long-range dependency modeling.
    \item 
    Experimental results on both text and image tasks demonstrate that MixFormer achieves strong performance while supporting efficient parallel training and fast autoregressive inference, providing a promising solution for building efficient Transformer models.
\end{itemize}

\begin{figure}[thb]
\centering
\includegraphics[width=0.5\linewidth]{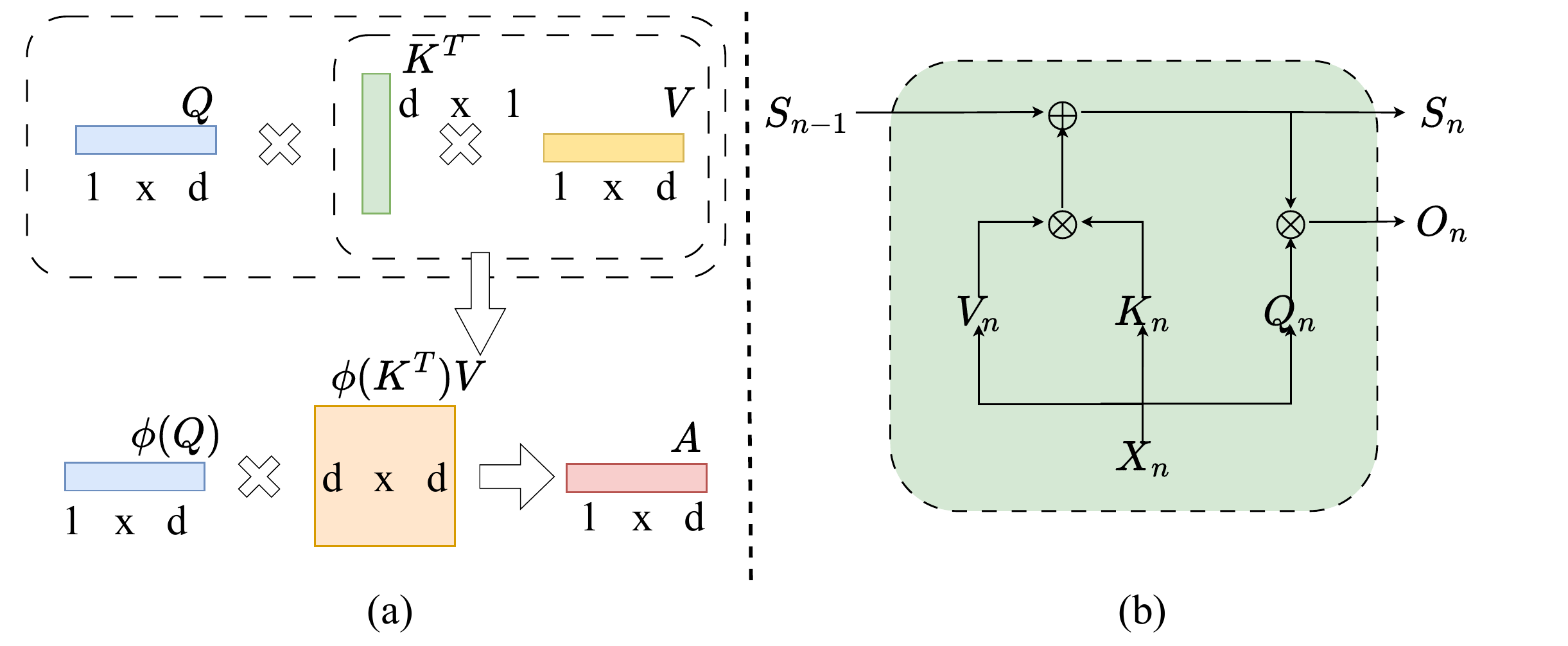}
    \caption{
    \textbf{Linear Attention vs SSMs.}
    (a) is the workflow of the Linear Attention, and (b) is the State Space Models.
    \label{fig:LAvsSSM}
    }
\end{figure}

\begin{figure*}[htb]
\centering
\includegraphics[width=0.8\linewidth]{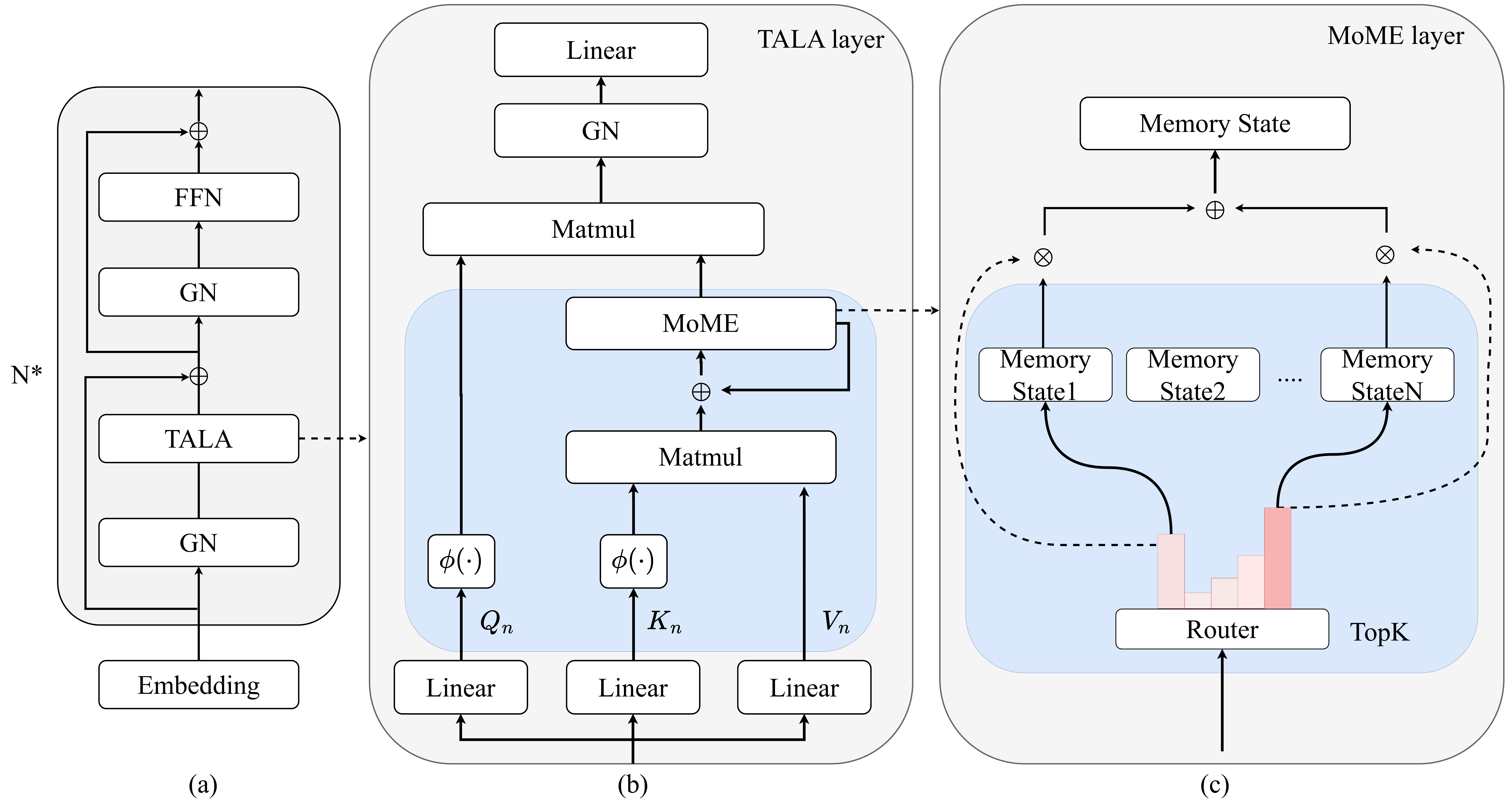}
    \caption{
    (a) is the overall architecture of our proposed MixFormer, (b) is the Time-Aware Linear Attention (TALA) layer, and (c) is the Mixture of Memory Experts (MoME) layer.
    \label{fig:mothod}
    }
\end{figure*}

\section{Preliminaries}
\label{sec:preliminaries}
\begin{figure}[thb]
\centering
\includegraphics[width=0.5\linewidth]{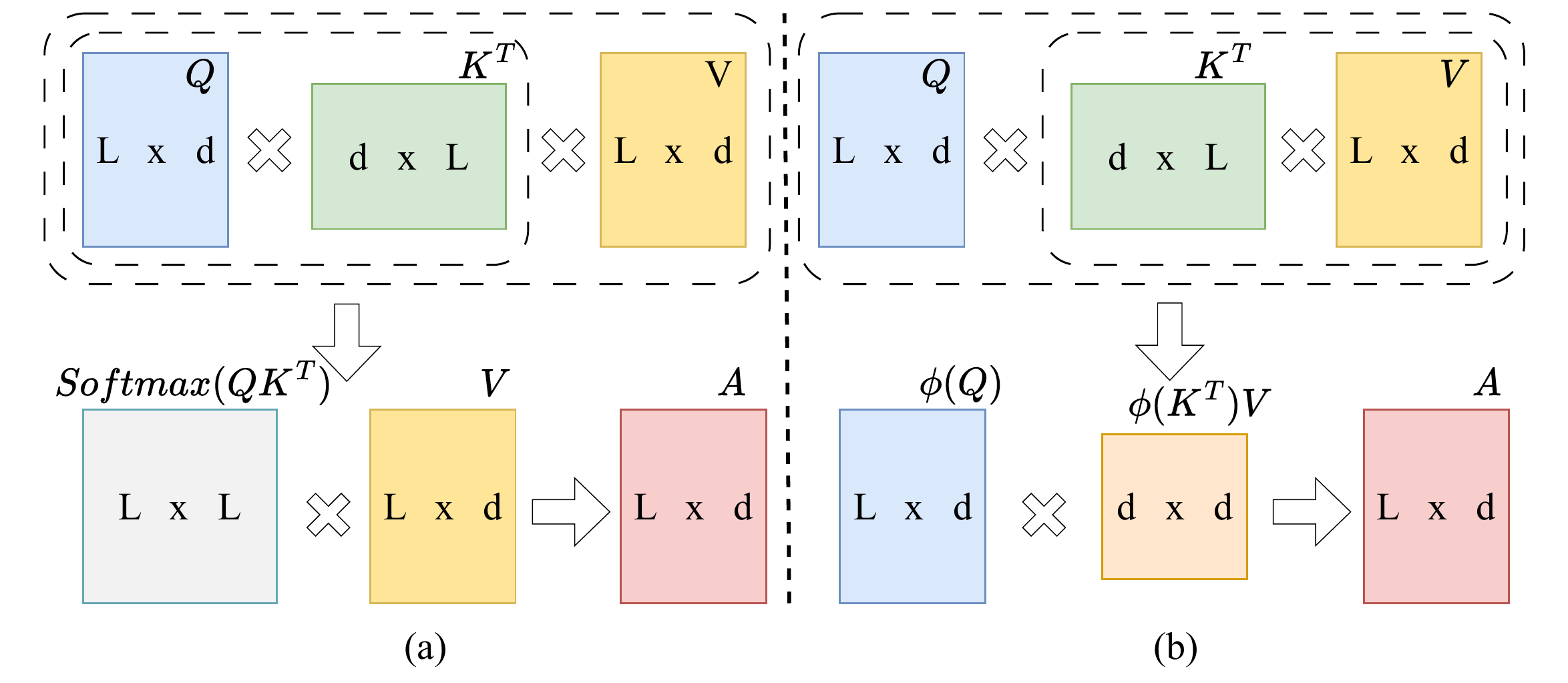}
    \caption{
    \textbf{Dot-product Attention vs Linear Attention.}
    (a) is the illustration of the Dot-product Attention, and (b) is the Linear Attention.
    \label{fig:compare_ATT}
    }
\end{figure}
To better understand the working details of our proposed MixFormer, we provide a detailed review of various forms of dot-product and linear attention.

\subsection{Dot-product Attention}
Figure \ref{fig:compare_ATT} (a) illustrates the dot-product attention mechanism in the Vanilla Transformer \cite{vaswani2017attention}.
Given the input $X$, it is mapped into three different semantic space matrices: $Q$, $K$, and $V$, where $Q, K, V \in R^{L\times d}$, $L$ denotes the context length, and $d$ is the hidden state dimension.
Next, the attention weight matrix between $Q$ and $K$ is computed using the softmax function.
Finally, this matrix is used to compute the weighted sum with V, the final result, as follows:

\begin{equation}
\begin{aligned}
\label{softmax_att}
 Q &= XW_q,\\K &= XW_k, \\V &= XW_v \\
Attention(Q,K,V) &= Softmax(\frac{QK^T}{\sqrt{d_k}})V,
\end{aligned}
\end{equation}
where $W_q$, $W_k$, and $W_v$ are the trainable matrix, $d_k$ is hidden layer dimensional number.
Generally, we refer to the attention mechanism as Dot-Product Attention or Softmax Attention.
Its time and space complexity are both quadratic with respect to the sequence length ($L$).

\subsection{Generalization of Attention}
According to Equation (\ref{softmax_att}), the attention vector for the \( i \)-th token is calculated as follows:
\begin{equation}
\label{softmax_att_i}
    Attention(Q,K,V)_i = \sum_{j=1}^{n}\frac{e^{q_i^T k_j}}{\sum_{j=1}^{n}e^{q_i^T k_j}}v_j,
\end{equation}
where $q$, $k$, and $v$ are column vectors.
The $e^{q_i^\top k_j}$ is used as a similarity function between two vectors.
Let the similarity function be denoted as \( sim(q_i, k_j) \).
This generalizes Equation (\ref{softmax_att_i}) to Equation (\ref{sim_att_i}), following:
\begin{equation}
\label{sim_att_i}
    Attention(Q,K,V)_i = \sum_{j=1}^{n}\frac{sim(q_i, k_j)}{\sum_{j=1}^{n}sim(q_i, k_j)}v_j,
\end{equation}
where $sim(q_i, k_j) >= 0$,
following previous work \cite{katharopoulos2020transformers, shen2021efficient}, the similarity function must ensure the non-negativity of the attention map.

\subsection{Linear Attention}
The dot-product attention uses the softmax function to compute the similarity between $Q$ and $K$, resulting in a time and space complexity of \( O(N^2) \).
Therefore, reducing the similarity function's time and space complexity is key to achieving an efficient transformer.
Figure \ref{fig:compare_ATT} (b) illustrates linear attention, which uses a carefully designed similarity function, reducing the time and space complexity from $O(N^2)$ to $O(N)$.
Many previous works further extend $sim(\cdot, \cdot)$ to kernel-based methods \cite{schlag2021linear,han2023flatten,mengpolaformer}, which approximate the softmax operation.
The formula is as follows:

\begin{equation}
sim(q_i, k_j) = \phi(q_i)^T \phi(k_j),
\end{equation}
where $\phi(\cdot)$ can be seen as kernel smoother.

By substituting the kernel similarity function into Equation (\ref{sim_att_i}), we obtain:
\begin{equation}
    Attention(Q,K,V)_i = \sum_{j=1}^{n}\frac{\phi(q_i)^T \phi(k_j)}{\sum_{j=1}^{n}\phi(q_i)^T \phi(k_j)}v_j
\end{equation}

Then, by exploiting the associativity of matrix multiplication, the computation order of $q$, $k$, and $v$ is altered.
It can be simplified to:
\begin{equation}
    Attention(Q,K,V)_i = \frac{\phi(q_i)^T\sum_{j=1}^{n} (\phi(k_j)v_j)}{\phi(q_i)^T\sum_{j=1}^{n} \phi(k_j)}
\end{equation}

The intermediate results are reduced from an $L\times L$ matrix to a $d\times d$ matrix, and the overall computational complexity becomes $\mathcal{O}(d^2L)$, which scales linearly with the sequence length $L$.

\subsection{Autoregressive Generation}
In the inference process of Linear Transformers, the prediction of the current token depends solely on the previously generated token sequence and is not influenced by any subsequent tokens.
Following the autoregressive generation formulation of language models in \cite{katharopoulos2020transformers}, we reformulate Equation (\ref{sim_att_i}) into an autoregressive computation paradigm:
\begin{equation}
    Attention(Q,K,V)_i = \frac{\phi(q_i)^TS_i}{\phi(q_i)^TZ_i},
\end{equation}
where $S_i$ and $Z_i$ are two newly defined variables, as follows:
\begin{equation}
\begin{aligned}
    &S_i = \sum_{j=1}^{i} (\phi(k_j)v_j),Z_i = \sum_{j=1}^{i} \phi(k_j) \\
    &S_i = S_{i-1}+\phi(k_j)v_j^T\\
    &Z_i = Z_{i-1}+\phi(k_j),\\
\end{aligned}
\end{equation}
where $S_0 = 0$ and $Z_0 = 0$.
Therefore, the linear attention mechanism enables efficient autoregressive inference with both linear time complexity and constant memory usage.




\section{Related Work}
\subsection{Linear Transformers}

Linear Transformers represent one of the main directions in the development of efficient Transformers.
As shown in Figure \ref{fig:LAvsSSM} (a), many studies aim to approximate dot-product attention through carefully designed kernel function variants, thereby changing the computation order of $QKV$ and reducing the cost complexity from $O(L^2)$ to $O(L)$.
Such designs enable linear Transformers to achieve performance comparable to that of vanilla Transformers \cite{vaswani2017attention}.
Li et al.\cite{li2007linear} proposed a Taylor approximation-based linear attention mechanism.
This method approximates the softmax function using the first-order Taylor expansion of the $e^x$ term in the softmax function.
Katharopoulos et al. \cite{katharopoulos2020transformers} proposed replacing kernel methods with non-negative activation functions, achieving up to 4000× faster autoregressive prediction on very long sequences.
Shen et al. \cite{shen2021efficient} proposed a method that ingeniously applies the softmax function separately to the rows of $Q$ and the columns of $K$.
Peng et al. \cite{peng2021random} proposed random feature methods that offer a straightforward way of learning with recency bias through an optional gating mechanism to approximate the softmax function.
Choromanski et al. \cite{choromanskirethinking} introduced Performers, which employ a novel fast attention via positive orthogonal random characteristics (FAVOR) approach to go beyond the limitations of softmax-based attention.
Han et al. \cite{han2023flatten} proposed a novel focusing function that makes similar vectors more concentrated and dissimilar vectors more dispersed.
Meng et al. \cite{mengpolaformer} proposed POLAFormer, which improves the focusing function by making its focusing capability learnable, thereby further enhancing model performance.

These efforts have advanced the development of linear Transformers.
However, kernel-based approximation methods often yield attention maps that fail to adequately focus on important features.
On the other hand, these methods fail to account for the recursive formulation of linear attention, thereby overlooking the update mechanism of the state memory matrix during inference.
In contrast, our proposed MixFormer integrates both the parallel and recurrent forms of linear attention, while further enhancing the dynamic temporal awareness of the state memory matrix.

\subsection{State Space Models}

Another line of research reformulates the computation paradigm of linear Transformers into an iterative process, enabling parallel modeling while maintaining autoregressive inference.
As shown in Figure \ref{fig:LAvsSSM} (b), these methods draw inspiration from the autoregressive nature of recurrent neural nets (RNNs) and can be regarded as RNN-style Transformers.
Such architectures are typically categorized as State Space Models (SSMs) \cite{guefficiently,gumamba,dao2024transformers}.
Existing studies have proposed various algorithmic improvements to SSMs, aiming to optimize the update mechanism of their historical information feature matrices.
Han et al. \cite{sun2023retentive} proposed the Retentive Network (RetNet), which uses a fixed parameter $\gamma \in [0,1]$ as a computational factor to update historical information in the SSM.
Han et al. \cite{sun2023retentive} proposed the Retentive Network (RetNet), which introduces a fixed parameter $\gamma \in [0,1]$ as a computational factor to update historical information in the SSM.
RetNet demonstrates the advantage of SSMs over kernel-based linear Transformers, enabling both Parallel training and sequential inference.
Meanwhile, it employs chunkwise Parallelism to model ultra-long sequences, where inputs are processed in Parallel within chunks and sequentially across chunks.
Yang et al. \cite{yang2024gated} further extended this idea with Gated Linear Attention (GLA), where a learnable weight matrix ($W$) is employed to control the update of historical information at each time step.
GLA uses an input-dependent forgetting gate and dynamically adjusts the update weights of the sequence.
Others of the SSM focus on dynamically updating the memory state at each time step through an exponential decay mechanism.
Zhai et al. \cite{zhai2021attention} proposed a learned pair-wise positional bias based on exponential decay, enabling SSMs to achieve dynamic temporal awareness.
Peng et al. \cite{peng2023rwkv} introduced RWKV, which leverages token-shifted weighted exponential decay to further optimize the memory modeling mechanism.
As the model depth increases, RWKV empowers SSMs with long-range temporal dependency modeling capabilities.
Gu and Dao \cite{gumamba} proposed the Mamba model, which designs the parameters of the SSM as functions of the input, thereby enabling dynamic control of information flow.
Subsequently, Dao and Gu \cite{dao2024transformers} introduced Mamba-2, an improved version of Mamba based on the State Space Duality (SSD) framework, which enhances inference efficiency.

However, merely enhancing the representational capacity of the memory state in SSMs is insufficient to address the problem of knowledge forgetting in long sequences.
Our proposed MixFormer introduces a Mixture-of-Memory-Experts mechanism, deploying multiple functionally specialized memory experts to capture different types and scales of temporal knowledge. This multi-expert collaborative architecture significantly improves the model’s ability to retain information over long-range dependencies, effectively mitigating progressive knowledge decay in long-sequence tasks.

\section{Methodology}
\subsection{Problem Definition}
As shown in Figure \ref{fig:mothod} (a), based on the Transformer architecture \cite{vaswani2017attention}, our proposed MixFormer adopts a similar multi-block stacking structure.
Each MixFormer block consists of a Time-Aware Linear Attention (TALA) and a Feed-Forward Network (FFN) layer, each followed by a Group Norm (GN) layer.
Given an input sequence $X = (x_1, \dots, x_n)$, where $n$ denotes the context length, MixFormer employs parallel encoding during training and autoregressive generation during inference.
The input vectors $X$ are first encoded through an embedding layer into the initial representations $X^0 = (x_1^0, \dots, x_n^0) \in R^{n \times d}$, where $d$ denotes the hidden dimension.
These representations are then iteratively refined through $L$ stacked MixFormer blocks as follows:
\begin{equation}
    X^l = MixFormer\_Block_l(X^{l-1}), \quad l \in [1, L]
    \label{eq:mix_block}
\end{equation}
where $l$ denotes the index of the current block layer, and $L$ is the total number of layers in the model.
Each block progressively enhances the contextual representations.
$MixFormer\_Block(\cdot)$ denotes the mapping process of the current input within this module.

MixFormer adopts a pre-norm strategy to enhance the training stability during deep stacking, ensuring effective gradient flow across multiple layers.
Meanwhile, Group Norm (GN) \cite{wu2018group} is introduced to address potential statistical biases arising from chunkwise computation, further improving the model’s training stability and convergence performance.
The TALA module incorporates time-aware linear attention to effectively capture long-range temporal dependencies among tokens, while supporting both parallel training and recursive inference modes.
The notation used throughout the methodology is summarized in Table \ref{tab:symbols}.

\subsection{Symbol Definition}
\label{sec:symbol-definition}
\begin{table}[htb] 
\centering
\caption{\textbf{Illustration of mathematical symbols.}}
\label{tab:symbols}
\begin{tabular}{cc}
\toprule
\textbf{Symbol} & \textbf{Description} \\
\midrule
$X$ & Model inputs \\
$X^l$ & Input to the $l$-th MixFormer block \\
$Q$ & Query projection matrix \\
$K$ & Key projection matrix \\
$V$ & Value projection matrix \\
$W_Q, W_K, W_V$ & Learnable parameter matrices \\
$\omega$ & Exponentially decayed weighting \\
$G$ & Learnable pair-wise positional bias \\
$S_n$ & Intermediate memory state matrix \\
$O$ & Model output \\
$B$ & Chunk length \\
\bottomrule
\end{tabular}
\end{table}

\subsection{Time-Aware Linear Attention}
\begin{figure}[thb]
\centering
\includegraphics[width=0.5\linewidth]{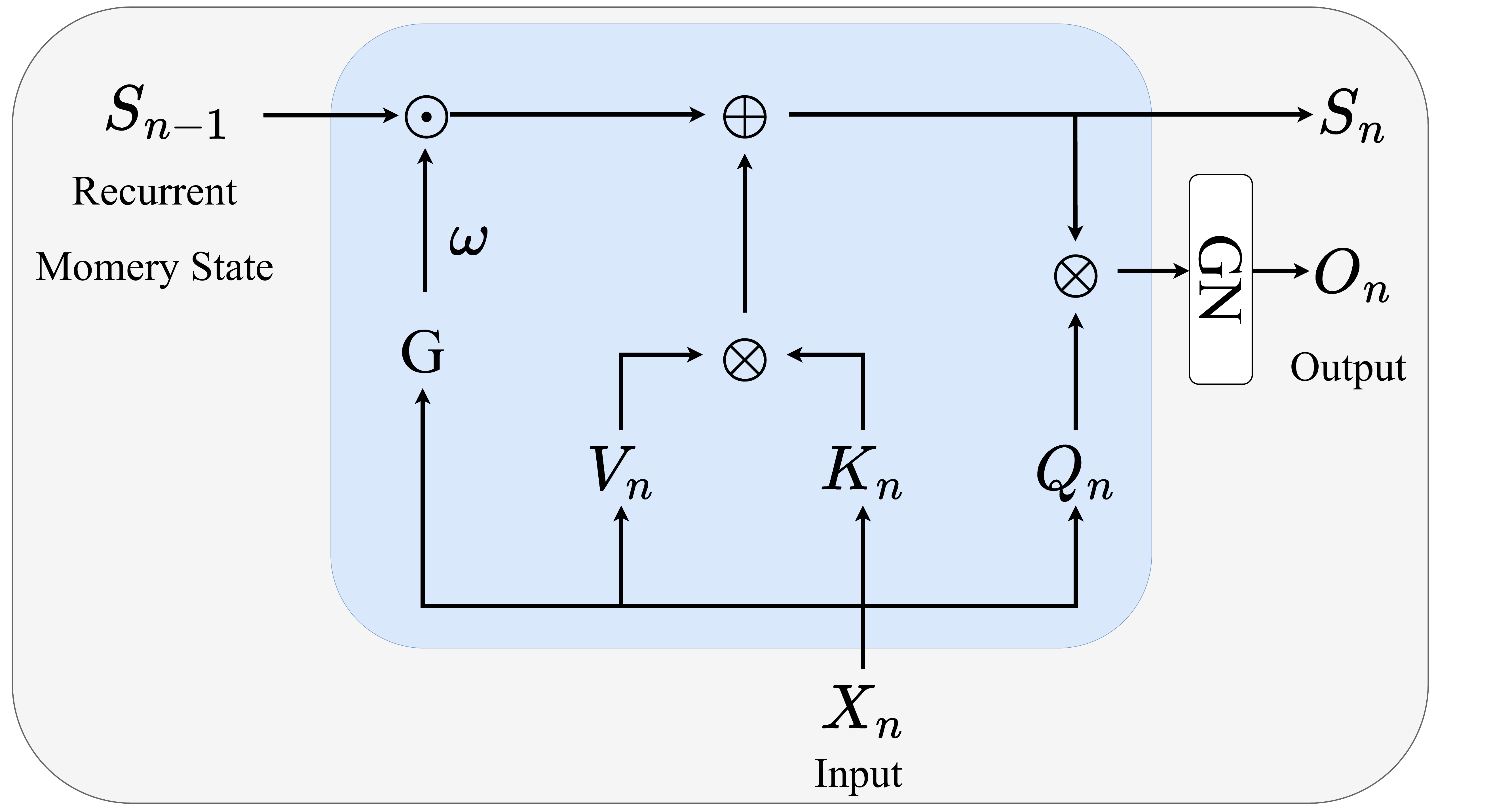}
    \caption{The recurrence of the Time-Aware Linear Attention.
    \label{fig:ssm}
    }
\end{figure}

To enhance the memory state perception of long sequences in the state space model, we propose a novel Time-Aware Linear Attention (TALA) layer to replace the standard multi-head self-attention layer.
Built upon the principles of linear attention and SSM design, this layer supports both parallel training and sequential inference.
The parallel computation structure is illustrated in Figure \ref{fig:mothod} (b), while the recursive computation process is shown in Figure \ref{fig:ssm}.

In the TALA layer, given an input sequence $X$, it is first projected into three vector spaces of the same dimension, as follows:
\begin{equation}
\begin{aligned}
    Q = XW_Q, K = XW_K, V = XW_V,\\
\end{aligned}
\end{equation}
where $W_Q \in {R}^{d \times d}$, $W_K \in {R}^{d \times d}$, and $W_V \in {R}^{d \times d}$ are learnable parameter matrices used to generate the query, key, and value vectors, respectively.
$d$ is the hidden dimension.

As shown in Section \ref{sec:preliminaries}, linear attention changes the computation order of $QKV$ and requires the non-negativity of $Q$ and $K$.
Following previous studies \cite{cai2022efficientvit, lu2021soft, xiong2021nystromformer}, TALA applies a kernel function transformation to ensure the non-negativity of $Q$ and $K$, which is defined as follows:
\begin{equation}
\begin{aligned}
&\phi(Q) = e^\frac{Q}{||Q||},\phi(K) = e^\frac{K}{||K||},\\
\end{aligned}
\label{fun_kel}
\end{equation}
where $|| \cdot ||$ denotes the $L_2$ normalization applied to the row vectors of $Q$ and the column vectors of $K$.

Meanwhile, we design a hybrid mechanism that integrates positional awareness with dimension-wise adaptability to enhance the temporal awareness of key–value (KV) pairs.
This mechanism applies exponentially decayed weighting based on relative positions to emphasize recent information, and employs a learnable parameter matrix to achieve adaptive calibration across feature dimensions.
While ensuring the non-negativity of $K$, it significantly improves the model’s ability to capture temporal dependencies in long sequences. The detailed computation process is as follows:

\begin{equation}
\begin{aligned}
&TALAtt(Q_t,K_t,V_t) = \phi(Q_t) \odot \frac{\sum^{t}_{i=1}\phi(K_i)(\omega G_{i})^{t-i}V_i}{\sum^{t}_{i=1}\phi(K_i)(\omega G_{i})^{t-i}}, \\
\label{TALA-eq}
\end{aligned}
\end{equation}
where $\odot$ denotes element-wise multiplication, $\omega$ represents exponentially decayed weighting, and $G \in R^{T*T}$ is a learnable pair-wise positional bias.
$TALAtt(\cdot)$ denotes the time-aware linear attention.

In addition, TALA can be extended to a multi-head attention mechanism to enhance its feature extraction capability.
By splitting the hidden dimension $d$ into multiple head dimensions $d_{head}$ and assigning independent decay parameters $\omega$ to each attention head, the model can capture temporal dynamics across multiple scales.
Based on this, Equation \ref{TALA-eq} can be extended to the following multi-head time-aware linear attention formulation:

\begin{equation}
\begin{aligned}
    & \omega = 1-e^{-h}, h\in [1,\cdots,d/d_{head}] \\
    & head_i = TALAtt(Q_t,K_t,V_t,\omega_i), \\
    & Y = GroupNorm_h(Concat(head_1,\cdots,head_h)),
\end{aligned}
\end{equation}
where $d$ is hidden dimension, $d_{head}$ is head dimension.
The GroupNorm is applied to the output of each head.

\noindent \textbf{The Parallel Representation of TALA.}
During training, MixFormer can leverage parallel matrix operations.
The parallel computation architecture of TALA is illustrated in Figure \ref{fig:mothod} (b), and its computations are defined as follows:
\begin{equation}
\begin{aligned}
    &Q = \phi(XW_Q), K = \phi(XW_K), V = XW_V,\\
    &TALAtt(Q,K,V) = Q\odot(K^T \odot D)V , \\
    &D=
    \left\{
    \begin{array}{l}
    \omega^{n-m} G_{nm}, n \geq m\\
    0,\ n < m, \\
    \end{array}
    \right.\\
    \end{aligned}
    \label{eq:mmr}
\end{equation}
where $W_Q \in R^{d \times d}$, $W_K \in R^{d \times d}$, and $W_V \in R^{d \times d}$ are learnable parameter matrices, $D$ is a causal masking matrix, and $G_{nm}$ is a learnable pair-wise positional bias.

It can be seen that the proposed MoFormer reduces computational complexity by leveraging a linear attention mechanism, while also enabling efficient Parallel training on GPUs.

\noindent \textbf{The Recurrent Representation of TALA.}
During inference, MixFormer can reformulate TALA as a recurrent computation, enabling autoregressive inference.
This allows GPU memory consumption to remain constant, rather than growing with sequence length.
The recurrent computation structure of TALA is illustrated in Figure \ref{fig:ssm}, at the i-th time step, the computation proceeds through the following process:

\begin{equation}
\begin{aligned}
& S_0 = 0,\\
&S_n = \omega G\odot S_{n-1}+K_n^TV_n, \\
&O_n = Q_nS_n,\\
\end{aligned}
\end{equation}
where $G$ is the same as in Equation (\ref{eq:mmr}).
The TALA maintains linear-time complexity during inference.


\noindent \textbf{The Chunkwise Recurrent Representation of TALA.}
Chunkwise computation, an efficient training paradigm for long sequences, achieves significant acceleration by combining the advantages of parallel and sequential computation \cite{sun2023retentive, yang2024gated}.
In this approach, the input sequence is divided into consecutive chunks.
For each chunk, fully parallelized computation is performed to leverage hardware acceleration, while a recurrent state propagation mechanism is established across chunks to preserve long-range dependencies.
Under the linear attention framework, this hybrid computation structure not only enables faster training compared to conventional methods but also exhibits notable I/O-aware benefits through optimized memory access.
Specifically, let $B$ denote the chunk length. The computation of the $i$-th chunk in TALA is defined as follows:
\begin{equation}
\begin{aligned}
    & R_i = (\omega G)^B\odot R_{n-1}+K_{[i]}^TV_{[i]},\\
    & TALAtt(X_{[i]}) = \underbrace{Q_{[i]}\odot(K_{[i]}^T \odot D)V_{[i]}}_{inner-chunk} + \underbrace{Q_{[i]}R_{i-1}}_{cross-chunk},
\end{aligned}
\end{equation}
where $[i]$ is the $i$-th chunk.

\subsection{Mixture of Memory Experts}
In State Space Models (SSMs), a single memory state architecture suffers from inherent limitations in long-sequence reasoning: as the sequence length increases, information from early inputs inevitably decays.
While increasing the memory state dimension can partially mitigate this issue, it fundamentally fails to address the challenge of information forgetting under long-range dependencies.
We propose a Mixture-of-Memory-Experts (MoME) mechanism based on the Mixture-of-Experts (MoE) framework \cite{shazeer2017outrageously, narayan2018don} to overcome this problem.
This mechanism deploys multiple functionally heterogeneous memory expert networks, endowing the model with dynamic and differentiated memory capabilities: each expert specializes in memory patterns of varying temporal spans and information importance, and the system adaptively activates the most relevant subset of experts according to the current contextual state.
This design not only significantly enhances the model’s ability to recall historical information but also achieves long-term retention of critical information through a multi-granularity memory preservation strategy.

The MoME mechanism in TALA extends the memory state by maintaining multiple memory experts.
As shown in Figure \ref{fig:mothod} (c), at each time step, the system dynamically activates a subset of experts based on the current input and constructs the current memory state by weighted aggregation of their outputs. This mechanism significantly enhances the model’s ability to capture temporal features and salient information.
The computation is formally defined as follows:
\begin{equation}
\begin{aligned}
& Y_{ms}=\sum_{i=1}^{n}E_i MS_i,\\
\end{aligned}
\end{equation}
where $Y_{ms}$ denotes the weighted output of multiple memory experts, $MS_i$ represents the $i$-th memory expert, $E_i$ is a router weight vector that determines the contribution of each memory state expert, as follows:

\begin{equation}
\begin{aligned}
&E_i = Softmax(K_nV_nW_g),\\
&\sum_{j=1}^{k_e} E_i^j = 1,\\
\end{aligned}
\end{equation}
where $wg \in \mathbb{R}^{1 \times k_e}$ is a learnable parameter matrix, and $k_e$ is the number of experts.
The current time step’s $K_n$ is processed through the kernel transformation defined in Equation (\ref{fun_kel}).

Meanwhile, TALA enables the memory update at each time step to dynamically emphasize salient information while suppressing less relevant content.
With the stacking of multiple layers and the collaborative effect of the MoME mechanism, the model can effectively retain and utilize information from early tokens even after processing ultra-long sequences.

\subsection{Overall Architecture of MixFormer}
The proposed MixFormer consists of $N$ stacked MixFormer blocks (Equation (\ref{eq:mix_block})).
The input to the first block is the embedding of $X$, comprising both token and positional representations, while each subsequent block receives the output of the previous block as its input.
The computation is defined as follows:

\begin{equation}
\begin{aligned}
    &Y^l = TALA(GN(X^l)) + X^l, \\
    &X^{l+1} = FFN(GN(Y^l)) + Y^l, \\
\end{aligned}
\end{equation}
where $GN(\cdot)$ denotes the Group Normalization operation (GN).
MixFormer adopts a structure similar to that of mainstream Transformers and employs a Feed-Forward Network (FFN) to enable the modeling of complex nonlinear transformations.
The computation is defined as follows:
\begin{equation}
    FFN(X) = max(0, XW_1)W_2,
\end{equation}
where $W_1$ and $W_2$ are learnable weight matrices.

\subsection{Complexity Analysis}
To verify the linear-time complexity of MixFormer, we conduct a detailed complexity analysis as follows.
Let $N$ denote the sequence length (context length), $d$ the hidden state dimension, $d'$ the expansion dimension of the feed-forward network (FFN), where $d' > d$, and $k$ the number of memory experts.
The total computational cost of MixFormer comprises four main components: projections (Proj), the time-aware linear attention (TALA), mixture-of-memory experts (MoME), and the feed-forward network (FFN).
As shown in Equation~\ref{eq:complex}, the Projection cost (Query, Key, Value, and Output projections) is $4Nd^2$, the time-aware linear attention cost is $Nd^2$, the MoME cost is $Nk$, and the FFN cost is $2dd'$.
Under the long-context setting ($N \gg d$), the total computational complexity of MixFormer converges to $O(N)$.

\begin{equation}
     \underbrace{4Nd^2}_{Proj} + \underbrace{Nd^2}_{TALA} + \underbrace{Nk}_{MoME} + \underbrace{2dd'}_{FFN}
    \label{eq:complex}
\end{equation}

\subsection{Differences from Previous Methods}
To compare MixFormer with other efficient Linear Transformers, we conduct a comparative analysis across four key dimensions, as shown in Table \ref{tab:Dif_PM}.
For training parallelism, MixFormer, like the standard Transformer, supports fully parallelized training. In contrast, Recurrent Neural Networks (RNNs) and RWKV \cite{peng2023rwkv} suffer from sequential training bottlenecks. MixFormer achieves performance comparable to that of standard Transformers, while significantly outperforming Linear Transformers—whose kernel approximations often lead to degraded attention focus—as well as traditional RNN-based models.
For inference efficiency, MixFormer and linear Transformers both achieve $O(1)$ inference time per token, significantly outperforming the $O(N^2)$ complexity of standard Transformers.
For memory consumption, MixFormer maintains linear $O(N)$ memory usage with respect to sequence length, in contrast to the quadratic memory growth of standard Transformers.
MixFormer demonstrates significant advantages in its memory state mechanism.  Unlike conventional Transformers that rely on explicit attention score matrices and lack an inherent memory design, existing linear Transformers and SSMs typically maintain a single static memory state, which limits their ability to model complex dependencies over long sequences.  In contrast, MixFormer uses a mixture-of-expert memory mechanism, where a dynamic routing strategy assigns input features to multiple specialized memory modules, enabling the model to adaptively activate historical information across different dimensions.  This multi-granularity collaborative memory architecture allows MixFormer to capture fine-grained temporal patterns and retain information more effectively in long-sequence tasks.
Finally, compared to S4 \cite{guefficiently} and RetNet \cite{sun2023retentive}, MixFormer achieves stronger overall performance while preserving both training Parallelism and linear time and memory complexity. In summary, the proposed MixFormer effectively integrates the advantages of existing efficient Transformers and achieves significant performance improvements.

\begin{table*}[thb]
\centering
\caption{Comparison between MixFormer and other efficient linear Transformer models from multiple perspectives.
Including training parallelism, inference cost, memory consumption on long sequences, and the number of memory states.
}
\resizebox{\textwidth}{!}{%
\begin{tabular}{ccccc}
\toprule
    Architectures & Training
Parallelization &  Inference Cost & Long-Sequence
Memory Complexity & Memory State\\
    \midrule
    Transformer & Y & $O(N)$ &$O(N^2)$& N/a\\
    Linear Transformer & Y & $O(1)$ &$O(N)$& Singe\\
    Recurrent NN & N & $O(1)$ &$O(N)$& Singe\\
    RWKV & N & $O(1)$ &$O(N)$& Singe\\
    S4 & Y & $O(1)$ &$O(NlogN)$& Singe\\
    RetNet & Y & $O(1)$ &$O(N)$& Singe\\
    \midrule
    \textbf{MixFormer (Ours)} & Y & \textbf{$O(1)$} &\textbf{$O(N)$} & Multiple \\
    \bottomrule
\end{tabular}
}
\label{tab:Dif_PM}
\end{table*}

\section{Experiments}
\begin{table*}[thb]
\centering
\caption{
Comparison of MixFormer with other efficient Transformers on the LRA benchmark ($\%$).
The best results are highlighted in bold, and the second-best results are underlined.
The results of other models follow \cite{taylong}.
\label{tab:LRA}
}
\begin{tabular}{ccccccc}
\toprule
    Model & ListOps &  Text & Retrieval&Image&Pathfinder & Avg \\
    \midrule
    Transformer &36.37 &64.27 &57.46& 42.44& 71.40 & 54.39\\
    \midrule
    Local Attention &15.82 & 52.98 &53.39 &41.46 &66.63 &46.06\\
    Sparse Trans. &17.07&63.58& \underline{59.59}& 44.24& 71.71 &51.24\\
    Longformer &35.63&62.85 &56.89 &42.22 &69.71 & 53.46\\
    Linformer &35.70& 53.94& 52.27& 38.56 &76.34 &51.36\\
    Reformer &\underline{37.27} &56.10 &53.40 &38.07 &68.50 & 50.67\\
    Sinkhorn Trans. &33.67 &61.20 &53.83 &41.23 &67.45 &51.39\\
    Synthesizer & 36.99 &61.68& 54.67& 41.61 &69.45 &52.88\\
    BigBird &36.05 &64.02& 59.29 &40.83 &74.87 & \underline{55.01}\\
    Linear Trans. &16.13 &\underline{65.90} &53.09 &42.34 &75.30 &50.55\\
    Performer &18.01& 65.40& 53.82& \textbf{42.77} & \textbf{77.05} &51.41\\
    \midrule
    \textbf{MixFormer (Ours)} & \textbf{37.35} &\textbf{66.56} &\textbf{60.50}& \underline{42.65}& \underline{76.55}& \textbf{56.72}\\
    \bottomrule
\end{tabular}
\end{table*}

\subsection{Experimental Setups}

The main hyperparameter settings of MixFormer are summarized in Table \ref{tab:ES}.
In this experiment, MixFormer was configured with two parameter variants: A0.3B-2B and A1B-7B.
The A0.3B-2B model uses a hidden dimension of 1024 for the Q/K/V/O projections, a feed-forward network dimension of 896, 8 attention heads, 64 memory state experts, and 12 stacked MixFormer blocks.
For training, we adopt a learning rate of $1 \times 10^{-4}$ with the Adam optimizer, and train the model on a corpus containing 15 billion tokens.
The total number of parameters is 2B, of which 0.3B are active parameters.
The A1B-7B model uses a hidden dimension of 2048 for the Q/K/V/O projections, a feed-forward network dimension of 1024, 16 attention heads, 64 memory state experts, and 16 stacked MixFormer blocks.
For training, we adopt a learning rate of $1 \times 10^{-5}$ with the Adam optimizer, and train the model on a corpus containing 15 billion tokens.
The total number of parameters is 7B, of which 1B is an active parameter.
All experiments are conducted on four NVIDIA RTX 4090 GPUs.

\begin{table}[htb]
\centering
\caption{Experimental Setups.}
\begin{tabular}{ccc}
\toprule
    HyperParameters & A0.3B-2B& A1B-7B\\
    \midrule
    Hidden Dimension & 1024 & 2048\\
    FFN Dimension & 896 & 1024 \\
    Num of Heads & 8 & 16\\
    Num of Layers & 12 & 16\\
    Num of Memory Experts & 64 & 64\\
    LR & 1e-4 & 1e-5\\
    Training Tokens & 15B & 30B\\
    \bottomrule
\end{tabular}
\label{tab:ES}
\end{table}

\subsection{Long Context Ability}
To evaluate the effectiveness of MixFormer in long-context scenarios, we conduct experiments on the Long-Range Arena (LRA) benchmark \cite{taylong}, a standard suite for long-sequence modeling.
The Long Range Arena (LRA) benchmark systematically integrates five core tasks characterized by pronounced long-sequence dependencies. In the mathematical expression parsing task ListOps \cite{nangia2018listops}, models are required to parse deeply nested operator structures, posing a significant challenge to their hierarchical reasoning ability. The text classification task based on IMDb movie reviews \cite{maas2011learning} demands that models extract key sentiment features from lengthy reviews, testing their capacity for semantic filtering under noisy conditions. The AAN document retrieval task \cite{radev2013acl} evaluates how effectively models capture semantic associations across long academic documents through citation-based matching. The Pathfinder task \cite{linsley2018learning}, a synthetic visual challenge, requires the establishment of long-range spatial dependencies across pixel sequences. Finally, the CIFAR-10 image classification task \cite{krizhevsky2009learning}, which flattens images into sequences, innovatively examines a model’s ability to preserve visual features in the absence of two-dimensional structural priors.

The experimental results, as shown in Table \ref{tab:LRA}, demonstrate that MixFormer outperforms all compared efficient Transformer models in terms of overall average accuracy.
Specifically, it achieves the best performance on ListOps, text classification, and document retrieval tasks, and ranks second on CIFAR-10 and Pathfinder.
This can be attributed to MixFormer’s use of linear attention as the backbone, which enables longer sequence outputs under limited memory consumption, thereby enhancing contextual representation learning.
In addition, the MoME mechanism significantly broadens the model’s perceptual scope, enabling it to continuously focus on critical information in long-context scenarios.
By coordinating multiple specialized experts, this mechanism ensures that important content is effectively preserved and reinforced throughout long-sequence propagation, thereby enhancing the model’s ability to capture long-range dependencies.
Overall, MixFormer exhibits strong capability in modeling long-range dependencies across diverse domains.

\subsection{Image Generation}
\begin{figure}[htb]
\centering
\includegraphics[width=0.5\linewidth]{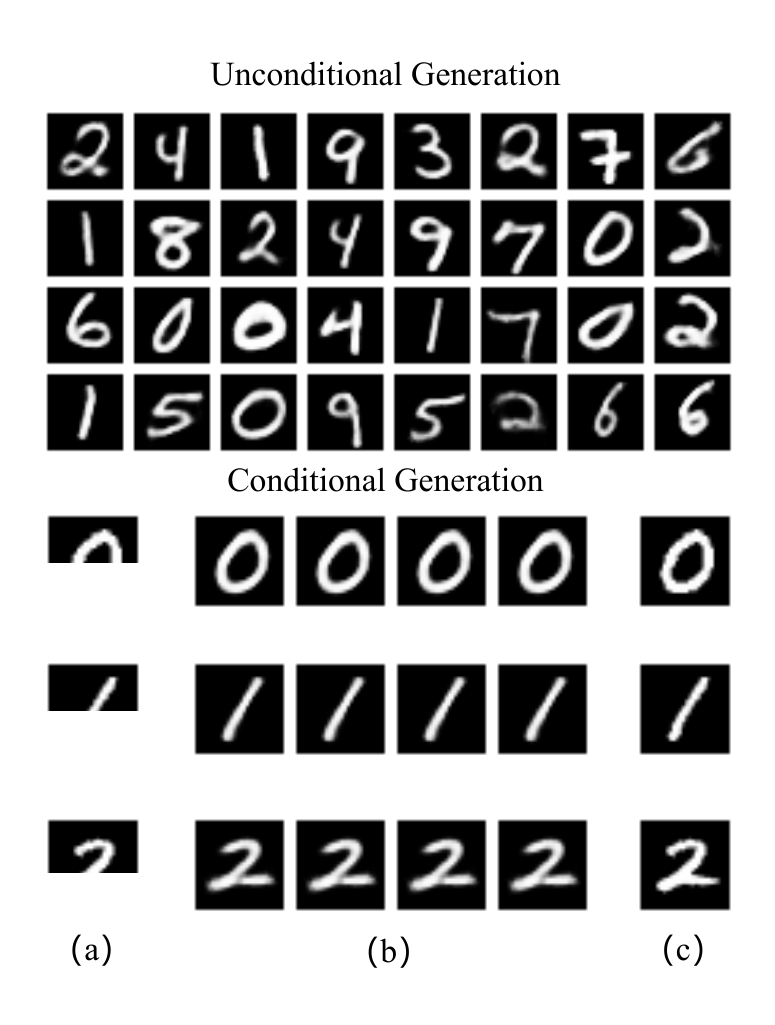}
\caption{
Experimental results on image generation tasks.
The top row is the result of unconditional generation, while the bottom row is conditional generation.
(a) is the occluded input image, (b) is the output generated by MixFormer, and (c) is the corresponding ground-truth image.
\label{fig:Image_Generation}
}
\end{figure}

To evaluate the effectiveness of MixFormer in image generation tasks, we conduct both conditional and unconditional generation experiments on the MNIST dataset \cite{cohen2017emnist}.
Figure \ref{fig:Image_Generation} (top row) is the result of unconditional generation.
While standard Transformers can be directly applied to image generation, they suffer from substantial memory overhead when modeling long sequences in pixel-by-pixel generation.
In contrast, MixFormer not only significantly reduces memory consumption and generation time but also produces high-quality samples with sharp boundaries and minimal noise.
Figure \ref{fig:Image_Generation} (bottom row) presents an image completion task, where the input consists of occluded images and MixFormer is responsible for generating the complete outputs.
We observe that MixFormer demonstrates strong long-range dependency modeling capabilities, accurately replicating the stroke style and width characteristics of the original images.
These advantages highlight the potential of MixFormer as an efficient and effective solution for long-sequence image generation tasks.
In summary, the proposed MixFormer performs remarkably well in both unconditional and conditional image generation tasks, demonstrating its effectiveness in modeling long-range pixel dependencies.

\subsection{CIFAR 10}
\begin{figure}[htb]
\centering
\includegraphics[width=0.5\linewidth]{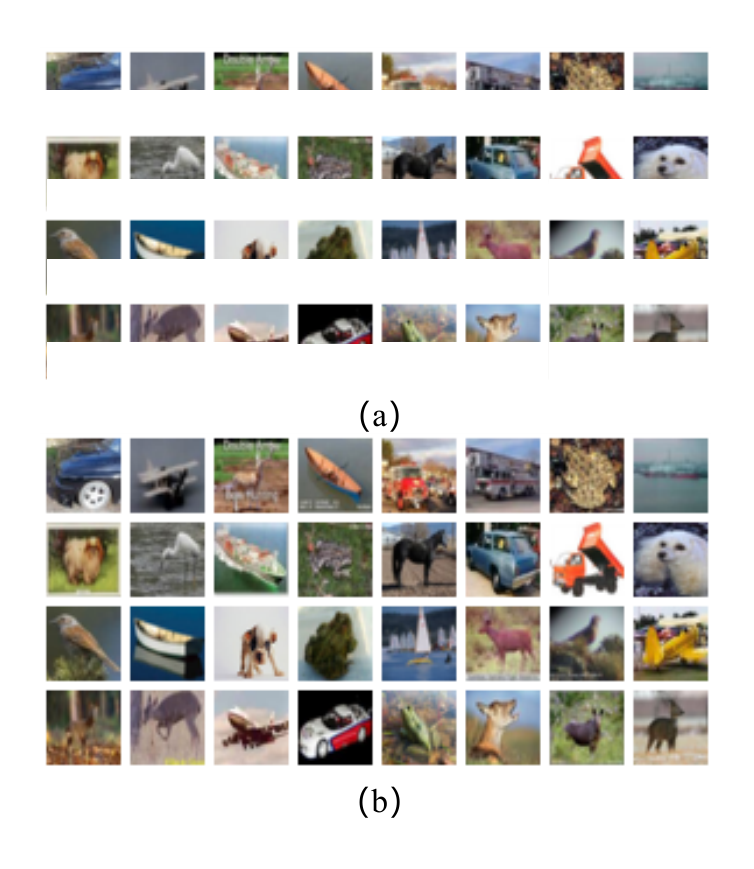}
\caption{
Experimental results on CIFAR 10.
(a) is the occluded input image, and (b) is the output generated by MixFormer.
\label{fig:cifar}
}
\end{figure}

To evaluate the performance of MixFormer on multi-channel image tasks, we conduct an image completion experiment on the CIFAR-10 \cite{krizhevsky2009learning}.
Compared with single-channel inputs, the serialized representation of multi-channel images results in substantially longer sequences, imposing greater demands on the model’s ability to capture long-range dependencies.
As shown in Figure \ref{fig:cifar}, when 50\% of image pixels are randomly masked, MixFormer successfully reconstructs visually coherent images with well-preserved structural integrity, clear boundaries, and minimal noise—significantly outperforming baseline models in texture continuity and detail preservation.
Within the MoME architecture, different memory experts specialize in capturing local texture features and global structural information. When parts of an image are occluded, the model adaptively activates the corresponding structural experts to recover the overall layout while engaging texture experts to refine fine-grained details.
In summary, the synergistic interaction between the MoME and the TALA mechanisms endows MixFormer with distinct advantages in multi-channel image completion tasks.

\subsection{Interpretability Analysis}
\begin{figure}[htb]
\centering
\includegraphics[width=0.5\linewidth]{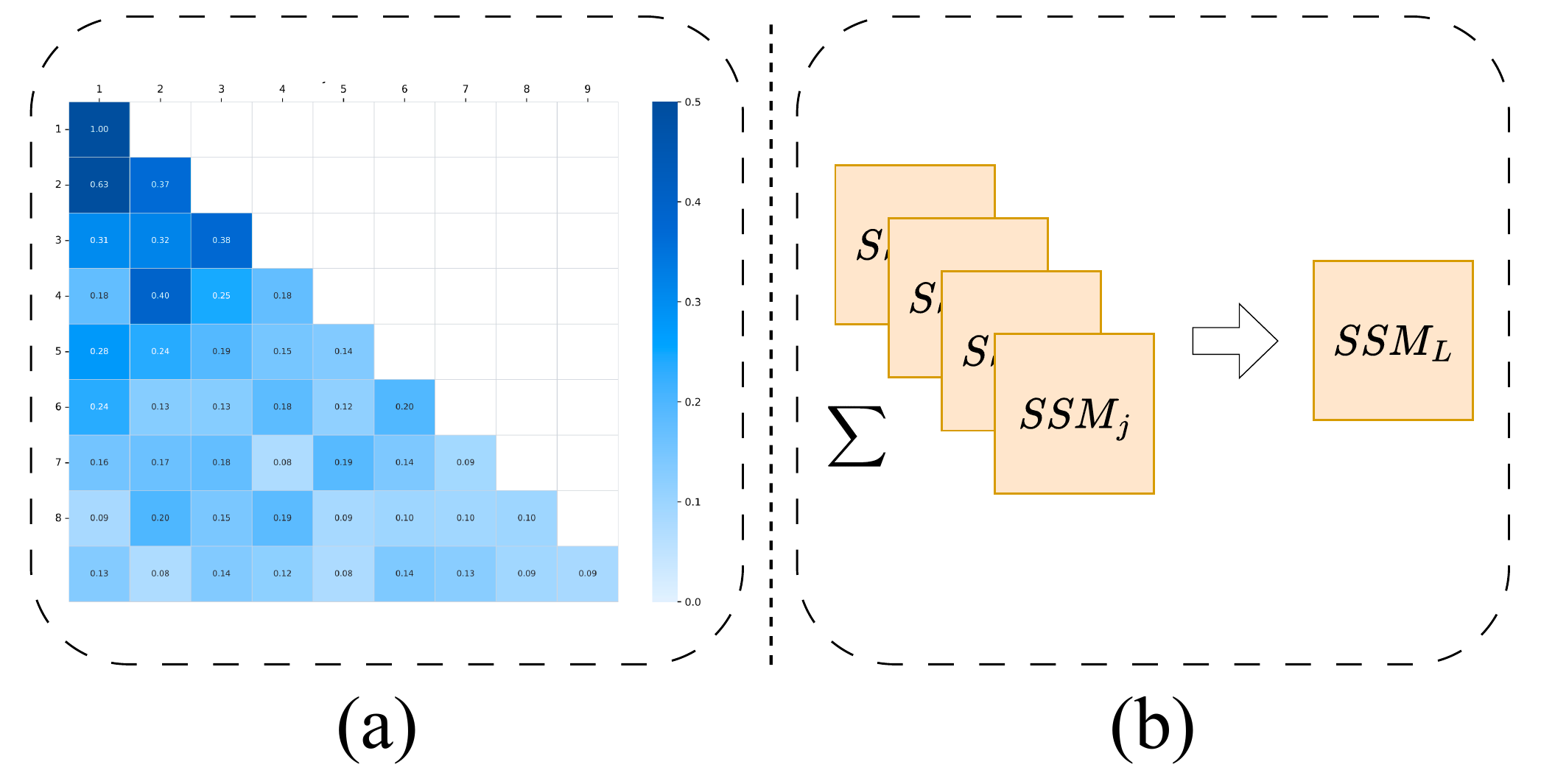}
\caption{
Visualized intermediate outputs of (a) standard dot-product attention and (b) reflection attention.
\label{fig:Interpretability_Analysis}
}
\end{figure}

We conduct an interpretability analysis by visualizing intermediate outputs between the standard dot-product attention and reflection attention.
Figure \ref{fig:Interpretability_Analysis} (a) shows the intermediate result of the standard softmax attention: the matrix $QK^T$ after softmax normalization.
Softmax attention not only strengthens the connections between important tokens but also suppresses irrelevant ones.
Moreover, it preserves full pairwise interactions among all tokens, ensuring comprehensive global information exchange.
Any given token can retrieve interaction information with all other tokens through the attention map.
In contrast, Figure \ref{fig:Interpretability_Analysis} (b) presents the $K^TV$ output in the reflection attention, which forms a fixed-size $d\times d$ matrix known as the State Space Model (SSM).
This matrix serves as a compressed global memory of historical information.
Due to its fixed dimensionality, the SSM does not retain explicit step-wise positional indices.
At each time step, the token updates the SSM by accumulating information from all previous steps.
At each time step, token representations are selectively written into the SSM based on the current input, and the output is generated using the updated global memory.

\begin{figure}[htb]
\centering
\includegraphics[width=0.5\linewidth]{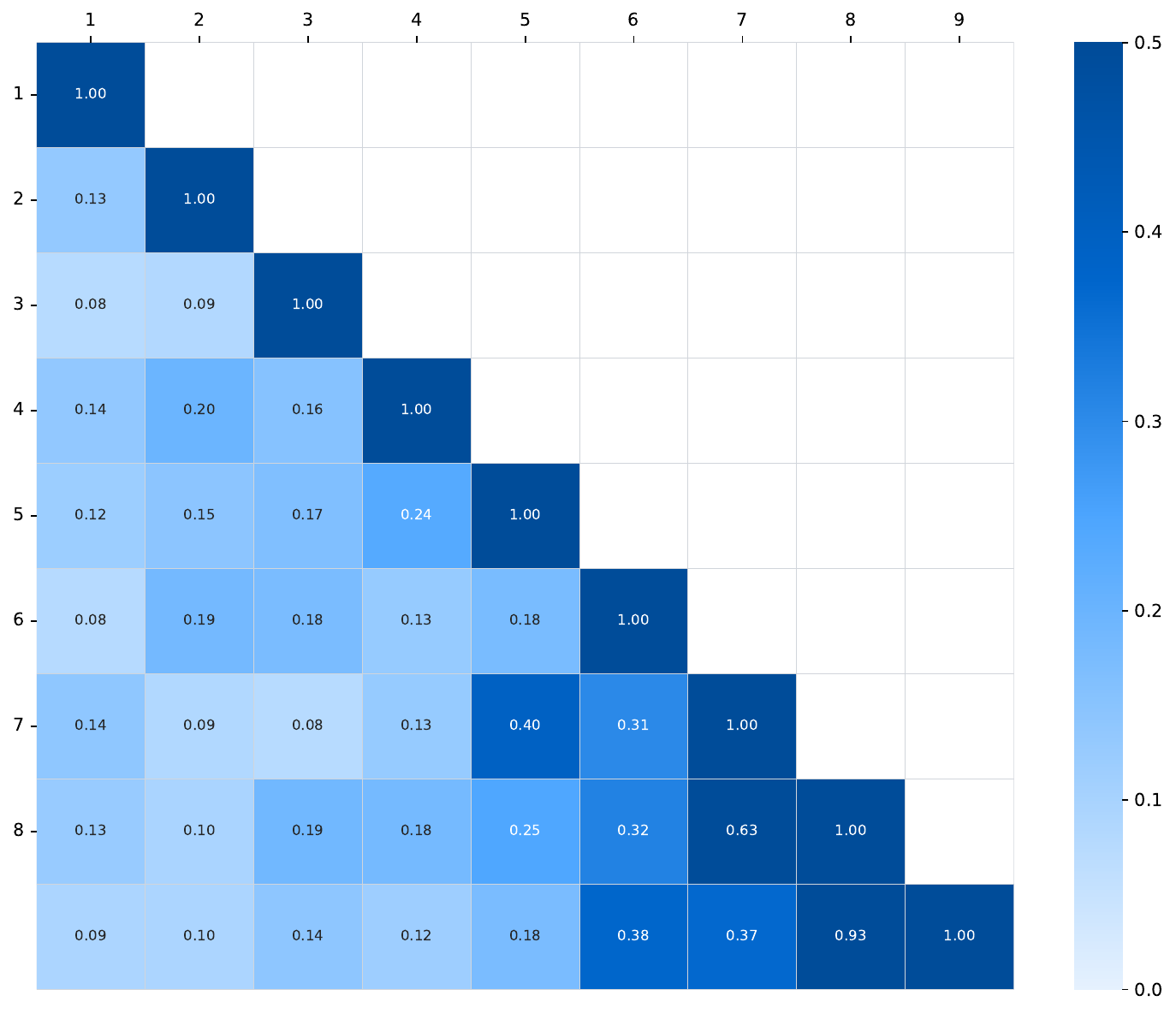}
\caption{
Visualization of the causal mask used by MixFormer.
\label{fig:MoME_inter}
}
\end{figure}

In other words, softmax attention produces explicit token-pair attention scores, while linear attention constructs an implicit global attention matrix.
During autoregressive inference, MixFormer enforces causal masking through the matrix $D$ (Equation \ref{eq:mmr}).
As illustrated in Figure \ref{fig:MoME_inter}, the visualization of the $D$ matrix clearly reveals the temporal evolution of memory attention values—an effect that arises from our exponential decay mechanism and learnable pairwise interaction biases.
In summary, the causal mask analysis demonstrates that MixFormer effectively balances strict causal constraints with global information flow.
Meanwhile, the multi-expert case study reveals a natural functional differentiation among memory experts—where some focus on short-term feature extraction while others specialize in long-term dependency preservation—offering a mechanistic explanation for MixFormer’s superior performance on long-sequence tasks.

\subsection{Mixture-of-Memory-Experts Analysis}
\begin{figure}[htb]
\centering
\includegraphics[width=0.5\linewidth]{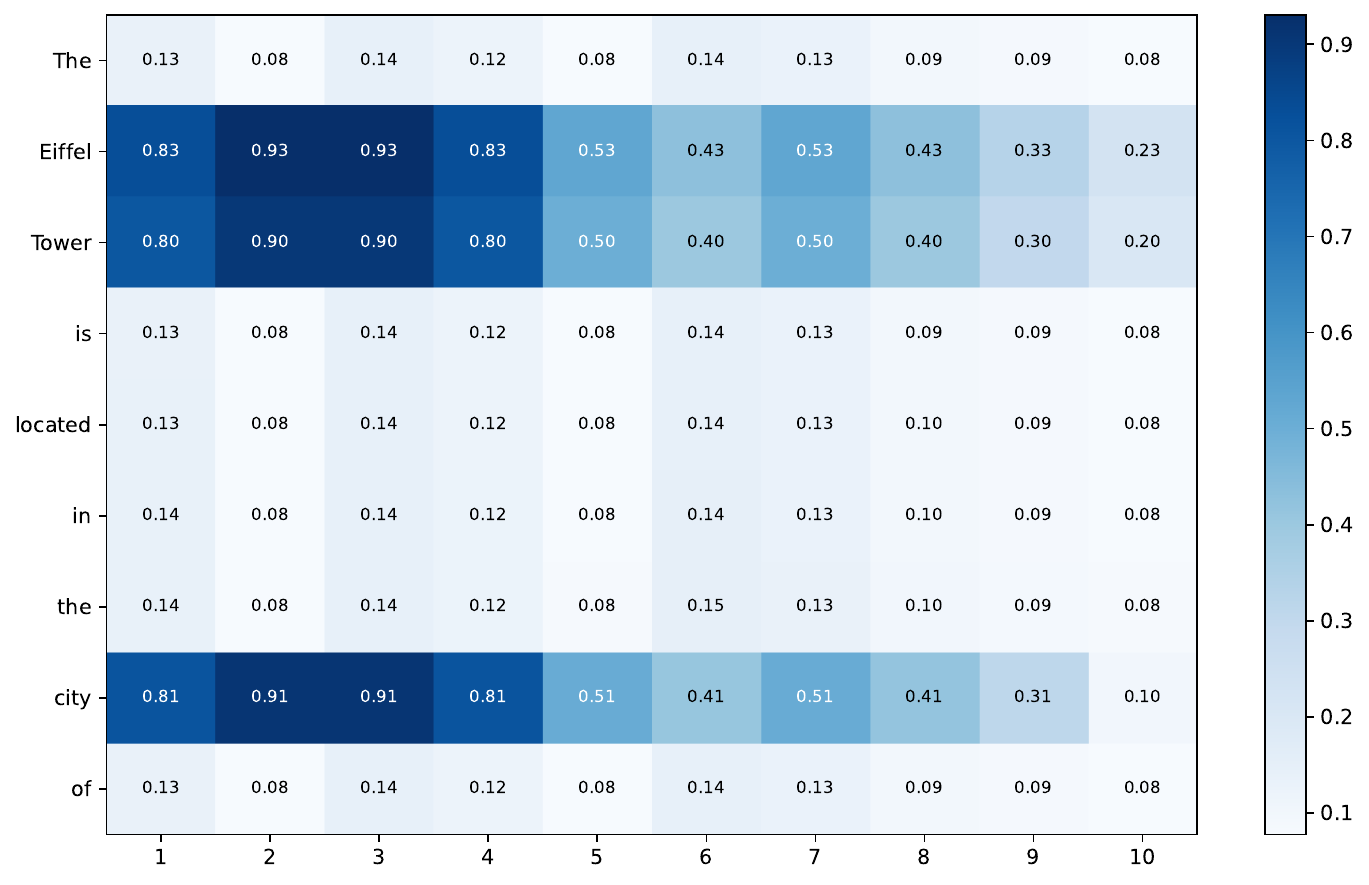}
\caption{
Visualization of memory-expert activations.
\label{fig:MoME_inter2}
}
\end{figure}

To examine the activation behavior of the MoME mechanism, we conduct a visualization analysis.
As shown in Figure \ref{fig:MoME_inter2}, when the current token is “Paris” and the context is “The Eiffel Tower is located in the city of”, the MoME mechanism exhibits significantly stronger expert activations for the tokens “Eiffel”, “Tower”, and “city”.
Through the collaboration of multiple memory experts, the model effectively identifies the strong semantic association between the landmark entity “Eiffel Tower” and “Paris”.
Meanwhile, the expert system captures the spatial hierarchical relationship between “city” and the target location “Paris”.
In particular, the long-term memory experts successfully preserve information about distant core entities, ensuring accurate contextual recall during generation.
This observation confirms that the MoME mechanism achieves dynamic selection and persistent retention of crucial historical information through functional specialization among experts.

\clearpage
\section{Conclusion}
In this work, we proposed MixFormer, a novel linear Transformer architecture incorporating the MoE to address the limitations of existing SSMs in ultra-long sequence modeling.
By maintaining multiple differentiated memory states in parallel and introducing the Time-Aware Linear Attention (TALA) mechanism, MixFormer dynamically updates memory with input-adaptive exponential decay and learnable positional biases.
This enables the model to selectively retain important historical information while mitigating memory dilution, effectively enhancing its capacity for long-range dependency modeling.
Extensive experiments on both long-sequence text and image generation tasks demonstrate that MixFormer consistently outperforms prior approaches, achieving robust performance under both parallel training and recurrent inference modes.    
Finally, MixFormer provides a promising solution for efficient Transformer design with strong modeling capabilities for ultra-long contexts.   
In future work, we plan to improve MixFormer in two directions. 
First, we aim to scale up the model parameters to further enhance its performance. 
Second, we will extend the application of MixFormer to a broader range of tasks and domains.

\bibliographystyle{unsrt}
\bibliography{mixformer}

@String{Computing = "Computing" }

@String{Computer = "{IEEE} Computer" }

@ArtifactSoftware{R,
    title = {R: A Language and Environment for Statistical Computing},
    author = {{R Core Team}},
    organization = {R Foundation for Statistical Computing},
    address = {Vienna, Austria},
    year = {2019},
    url = {https://www.R-project.org/},
}

@article{vaswani2017attention,
  title={Attention is all you need},
  author={Vaswani, Ashish and Shazeer, Noam and Parmar, Niki and Uszkoreit, Jakob and Jones, Llion and Gomez, Aidan N and Kaiser, {\L}ukasz and Polosukhin, Illia},
  journal={Proceedings of the 2017 Advances in Neural Information Processing Systems, {NeuraIPS}},
  pages={5998--6008},
  year={2017}
}

@article{li2007linear,
  title={Linear attention mechanism: an efficient attention for semantic segmentation.},
  author={Li, R and Su, J and Duan, C and Zheng, S},
  journal={arXiv preprint arXiv:2007.14902},
  year={2020}
}

@inproceedings{shen2021efficient,
  title={Efficient attention: attention with linear complexities},
  author={Shen, Zhuoran and Zhang, Mingyuan and Zhao, Haiyu and Yi, Shuai and Li, Hongsheng},
  booktitle={Proceedings of the 2021 International Conference on Computer Vision, {ICCV}},
  pages={3531--3539},
  year={2021}
}

@inproceedings{taylong,
  title={Long range arena: a benchmark for efficient transformers},
  author={Tay, Yi and Dehghani, Mostafa and Abnar, Samira and Shen, Yikang and Bahri, Dara and Pham, Philip and Rao, Jinfeng and Yang, Liu and Ruder, Sebastian and Metzler, Donald},
  booktitle={Proceedings of the 2020 International Conference on Learning Representations, {ICLR}},
  year={2020}
}

@article{tay2022efficienttransformerssurvey,
author = {Tay, Yi and Dehghani, Mostafa and Bahri, Dara and Metzler, Donald},
title = {Efficient Transformers: a survey},
volume = {55},
number = {6},
journal = {ACM Computing Survey},
year = {2022},
}

@inproceedings{katharopoulos2020transformers,
  title={Transformers are rnns: fast autoregressive transformers with linear attention},
  author={Katharopoulos, Angelos and Vyas, Apoorv and Pappas, Nikolaos and Fleuret, Fran{\c{c}}ois},
  booktitle={Proceedings of the 2020 International Conference on Machine Learning, {ICML}},
  pages={5156--5165},
  year={2020}
}

@inproceedings{han2023flatten,
  title={Flatten transformer: vision transformer using focused linear attention},
  author={Han, Dongchen and Pan, Xuran and Han, Yizeng and Song, Shiji and Huang, Gao},
  booktitle={Proceedings of the 2023 International Conference on Computer Vision, {ICCV}},
  pages={5961--5971},
  year={2023}
}

@inproceedings{peng2021random,
  title={Random feature attention},
  author={Peng, H and Pappas, N and Yogatama, D and Schwartz, R and Smith, N and Kong, L},
  booktitle={Proceedings of the 2021 International Conference on Learning Representations, {ICLR}},
  year={2021}
}

@inproceedings{choromanskirethinking,
  title={Rethinking attention with performers},
  author={Choromanski, Krzysztof Marcin and Likhosherstov, Valerii and Dohan, David and Song, Xingyou and Gane, Andreea and Sarlos, Tamas and Hawkins, Peter and Davis, Jared Quincy and Mohiuddin, Afroz and Kaiser, Lukasz and others},
  booktitle={Proceedings of the 2021 International Conference on Learning Representations, {ICLR}},
  year={2021}
}

@inproceedings{guefficiently,
  title={Efficiently modeling long sequences with structured state spaces},
  author={Gu, Albert and Goel, Karan and Re, Christopher},
  booktitle={Proceedings of the 2022 International Conference on Learning Representations, {ICLR}},
  year={2022}
}

@article{sun2023retentive,
  title={Retentive network: a successor to transformer for large language models},
  author={Sun, Yutao and Dong, Li and Huang, Shaohan and Ma, Shuming and Xia, Yuqing and Xue, Jilong and Wang, Jianyong and Wei, Furu},
  journal={arXiv preprint arXiv:2307.08621},
  year={2023}
}

@inproceedings{yang2024gated,
  title={Gated linear attention transformers with hardware-efficient training},
  author={Yang, Songlin and Wang, Bailin and Shen, Yikang and Panda, Rameswar and Kim, Yoon},
  booktitle={Proceedings of the 2024 International Conference on Machine Learning, {ICML}},
  pages={56501--56523},
  year={2024}
}

@inproceedings{gumamba,
  title={Mamba: linear-time sequence modeling with selective state spaces},
  author={Gu, Albert and Dao, Tri},
  booktitle={Proceedings of the 2024 International First Conference on Language Modeling, {CoLM}},
  year={2024},
}

@inproceedings{dao2024transformers,
  title={Transformers are SSMs: generalized models and efficient algorithms through structured state space duality},
  author={Dao, Tri and Gu, Albert},
  booktitle={Proceedings of the 2024 International Conference on Machine Learning, {ICML}},
  pages={10041--10071},
  year={2024},
}

@article{cai2022efficientvit,
  title={Efficientvit: enhanced linear attention for high-resolution low-computation visual recognition},
  author={Cai, Han and Gan, Chuang and Han, Song},
  journal={arXiv preprint arXiv:2205.14756},
  volume={3},
  number={1},
  year={2022}
}

@article{lu2021soft,
  title={Soft: softmax-free transformer with linear complexity},
  author={Lu, Jiachen and Yao, Jinghan and Zhang, Junge and Zhu, Xiatian and Xu, Hang and Gao, Weiguo and Xu, Chunjing and Xiang, Tao and Zhang, Li},
  journal={Proceedings of the 2021 Advances in Neural Information Processing Systems, {NeuraIPS}},
  volume={34},
  pages={21297--21309},
  year={2021}
}

@inproceedings{xiong2021nystromformer,
  title={Nystr{\"o}mformer: a nystr{\"o}m-based algorithm for approximating self-attention},
  author={Xiong, Yunyang and Zeng, Zhanpeng and Chakraborty, Rudrasis and Tan, Mingxing and Fung, Glenn and Li, Yin and Singh, Vikas},
  booktitle={Proceedings of the 2021 Conference on Artificial Intelligence, {AAAI}},
  number={16},
  pages={14138--14148},
  year={2021}
}

@inproceedings{shazeer2017outrageously,
  title={Outrageously large neural networks: the sparsely-gated mixture-of-experts layer},
  author={Shazeer, Noam and Mirhoseini, Azalia and Maziarz, Krzysztof and Davis, Andy and Le, Quoc and Hinton, Geoffrey and Dean, Jeff},
  booktitle={Proceedings of the 2017 International Conference on Learning Representations, {ICLR}},
  year={2017}
}

@inproceedings{narayan2018don,
  title={Don’t give me the details, just the summary! topic-aware convolutional neural networks for extreme summarization},
  author={Narayan, Shashi and Cohen, Shay B and Lapata, Mirella},
  booktitle={Proceedings of the 2018 Conference on Empirical Methods in Natural Language Processing, {EMNLP}},
  pages={1797--1807},
  year={2018}
}

@article{touvron2023llama,
  title={Llama: open and efficient foundation language models},
  author={Touvron, Hugo and Lavril, Thibaut and Izacard, Gautier and Martinet, Xavier and Lachaux, Marie-Anne and Lacroix, Timoth{\'e}e and Rozi{\`e}re, Baptiste and Goyal, Naman and Hambro, Eric and Azhar, Faisal and others},
  journal={arXiv preprint arXiv:2302.13971},
  year={2023}
}

@article{touvron2023llama2,
  title={Llama 2: open foundation and fine-tuned chat models},
  author={Touvron, Hugo and Martin, Louis and Stone, Kevin and Albert, Peter and Almahairi, Amjad and Babaei, Yasmine and Bashlykov, Nikolay and Batra, Soumya and Bhargava, Prajjwal and Bhosale, Shruti and others},
  journal={arXiv preprint arXiv:2307.09288},
  year={2023}
}

@article{dubey2024llama,
  title={The llama 3 herd of models},
  author={Dubey, Abhimanyu and Jauhri, Abhinav and Pandey, Abhinav and Kadian, Abhishek and Al-Dahle, Ahmad and Letman, Aiesha and Mathur, Akhil and Schelten, Alan and Yang, Amy and Fan, Angela and others},
  journal={arXiv e-prints},
  pages={arXiv--2407},
  year={2024}
}

@article{deitke2024molmo,
  title={Molmo and pixmo: open weights and open data for state-of-the-art multimodal models},
  author={Deitke, Matt and Clark, Christopher and Lee, Sangho and Tripathi, Rohun and Yang, Yue and Park, Jae Sung and Salehi, Mohammadreza and Muennighoff, Niklas and Lo, Kyle and Soldaini, Luca and others},
  journal={arXiv e-prints},
  pages={arXiv--2409},
  year={2024}
}

@article{dai2024nvlm,
  title={Nvlm: open frontier-class multimodal llms},
  author={Dai, Wenliang and Lee, Nayeon and Wang, Boxin and Yang, Zhuolin and Liu, Zihan and Barker, Jon and Rintamaki, Tuomas and Shoeybi, Mohammad and Catanzaro, Bryan and Ping, Wei},
  journal={arXiv preprint arXiv:2409.11402},
  year={2024}
}

@inproceedings{zhuang2023survey,
  title={A survey on efficient training of transformers},
  author={Zhuang, Bohan and Liu, Jing and Pan, Zizheng and He, Haoyu and Weng, Yuetian and Shen, Chunhua},
  booktitle={Proceedings of the 2023 International Joint Conference on Artificial Intelligence, {IJCAI}},
  pages={6823--6831},
  year={2023}
}

@article{papa2024survey,
  title={A survey on efficient vision transformers: algorithms, techniques, and performance benchmarking},
  author={Papa, Lorenzo and Russo, Paolo and Amerini, Irene and Zhou, Luping},
  journal={IEEE Transactions on Pattern Analysis and Machine Intelligence},
  volume={46},
  number={12},
  pages={7682--7700},
  year={2024}
}

@article{han2022survey,
  title={A survey on vision transformer},
  author={Han, Kai and Wang, Yunhe and Chen, Hanting and Chen, Xinghao and Guo, Jianyuan and Liu, Zhenhua and Tang, Yehui and Xiao, An and Xu, Chunjing and Xu, Yixing and others},
  journal={IEEE Transactions on Pattern Analysis and Machine Intelligence},
  volume={45},
  number={1},
  pages={87--110},
  year={2022}
}

@article{gupta2022diagonal,
  title={Diagonal state spaces are as effective as structured state spaces},
  author={Gupta, Ankit and Gu, Albert and Berant, Jonathan},
  journal={Proceedings of the 2022 Advances in Neural Information Processing Systems, {NeuraIPS}},
  volume={35},
  pages={22982--22994},
  year={2022}
}

@article{gu2021combining,
  title={Combining recurrent, convolutional, and continuous-time models with linear state space layers},
  author={Gu, Albert and Johnson, Isys and Goel, Karan and Saab, Khaled and Dao, Tri and Rudra, Atri and R{\'e}, Christopher},
  journal={Proceedings of the 2021 Advances in Neural Information Processing Systems, {NeuraIPS}},
  volume={34},
  pages={572--585},
  year={2021}
}

@inproceedings{hasaniliquid,
  title={Liquid structural state-space models},
  author={Hasani, Ramin and Lechner, Mathias and Wang, Tsun-Hsuan and Chahine, Makram and Amini, Alexander and Rus, Daniela},
  booktitle={Proceedings of the 2023 International Conference on Learning Representations, {ICLR}},
  year={2023}
}

@inproceedings{smith2023simplified,
  title={Simplified state space layers for sequence modeling},
  author={Smith, Jimmy TH and Warrington, Andrew and Linderman, Scott W},
  booktitle={Proceedings of the 2023 International Conference on Learning Representations, {ICLR}},
  year={2023}
}

@inproceedings{mengpolaformer,
  title={PolaFormer: polarity-aware linear attention for vision transformers},
  author={Meng, Weikang and Luo, Yadan and Li, Xin and Jiang, Dongmei and Zhang, Zheng},
  booktitle={Proceedings of the 2025 International Conference on Learning Representations, {ICLR}},
year={2025}
}

@inproceedings{schlag2021linear,
  title={Linear transformers are secretly fast weight programmers},
  author={Schlag, Imanol and Irie, Kazuki and Schmidhuber, J{\"u}rgen},
  booktitle={Proceedings of the 2021 International Conference on Machine Learning, {ICML}},
  pages={9355--9366},
  year={2021}
}

@article{linsley2018learning,
  title={Learning long-range spatial dependencies with horizontal gated recurrent units},
  author={Linsley, Drew and Kim, Junkyung and Veerabadran, Vijay and Windolf, Charles and Serre, Thomas},
  journal={Proceedings of the 2018 Advances in Neural Information Processing Systems, {NeuraIPS}},
  volume={31},
  year={2018}
}

@inproceedings{maas2011learning,
  title={Learning word vectors for sentiment analysis},
  author={Maas, Andrew and Daly, Raymond E and Pham, Peter T and Huang, Dan and Ng, Andrew Y and Potts, Christopher},
  booktitle={Proceedings of the 2011 Annual Meeting of the Association for Computational Linguistics: Human Language Technologies, {ACL}},
  pages={142--150},
  year={2011}
}

@article{radev2013acl,
  title={The ACL anthology network corpus},
  author={Radev, Dragomir R and Muthukrishnan, Pradeep and Qazvinian, Vahed and Abu-Jbara, Amjad},
  journal={Language Resources and Evaluation},
  volume={47},
  number={4},
  pages={919--944},
  year={2013}
}

@article{krizhevsky2009learning,
  title={Learning multiple layers of features from tiny images},
  author={Krizhevsky, Alex and Hinton, Geoffrey and others},
  year={2009}
}

@inproceedings{nangia2018listops,
  title={ListOps: a diagnostic dataset for latent tree learning},
  author={Nangia, Nikita and Bowman, Samuel},
  booktitle={Proceedings of the 2018 Annual Meeting of the Association for Computational Linguistics: Human Language Technologies, {ACL}},
  pages={92--99},
  year={2018}
}

@inproceedings{cohen2017emnist,
  title={EMNIST: Extending MNIST to handwritten letters},
  author={Cohen, Gregory and Afshar, Saeed and Tapson, Jonathan and Van Schaik, Andre},
  booktitle={2017 international joint conference on neural networks (IJCNN)},
  pages={2921--2926},
  year={2017},
  organization={IEEE}
}

@article{zhai2021attention,
  title={An attention free transformer},
  author={Zhai, Shuangfei and Talbott, Walter and Srivastava, Nitish and Huang, Chen and Goh, Hanlin and Zhang, Ruixiang and Susskind, Josh},
  journal={arXiv preprint arXiv:2105.14103},
  year={2021}
}

@inproceedings{peng2023rwkv,
  title={RWKV: Reinventing RNNs for the Transformer Era},
  author={Peng, Bo and Alcaide, Eric and Anthony, Quentin and Albalak, Alon and Arcadinho, Samuel and Biderman, Stella and Cao, Huanqi and Cheng, Xin and Chung, Michael and Derczynski, Leon and others},
  booktitle={Findings of the Association for Computational Linguistics: EMNLP 2023},
  pages={14048--14077},
  year={2023}
}

@inproceedings{wu2018group,
  title={Group normalization},
  author={Wu, Yuxin and He, Kaiming},
  booktitle={Proceedings of the European conference on computer vision (ECCV)},
  pages={3--19},
  year={2018}
}

\end{document}